\documentclass{article}

\usepackage{microtype}
\usepackage{graphicx}
\usepackage{tikz}
\usetikzlibrary{arrows.meta, shapes.geometric, calc, positioning}
\usepackage{booktabs}
\usepackage{stfloats}
\usepackage{cuted}
\usepackage{amsmath}
\usepackage{amssymb}
\usepackage{mathtools}
\usepackage{amsthm}
\usepackage{capt-of}
\usepackage{hyperref}
\usepackage[accepted]{icml2026}
\usepackage[capitalize,noabbrev]{cleveref}
\hypersetup{pdfnewwindow=true}
\newcommand{\best}[1]{\textbf{#1}}
\newcommand{\nativebest}[1]{\underline{#1}}
\newcommand{\bothbest}[1]{\textbf{\underline{#1}}}
\newcommand{\AppendixTableSetup}{\centering\small\setlength{\tabcolsep}{5pt}\renewcommand{\arraystretch}{1.08}}

\theoremstyle{plain}

\theoremstyle{definition}

\theoremstyle{remark}

\definecolor{editblue}{RGB}{30,75,255}
\definecolor{editred}{RGB}{255,50,30}
\definecolor{tealobj}{RGB}{0,175,165}
\definecolor{sphereblue}{RGB}{32,70,175}
\definecolor{spherebrown}{RGB}{128,78,35}
\definecolor{darkgreen}{RGB}{55,145,65}
\definecolor{cubefront}{RGB}{18,82,205}
\definecolor{cuberight}{RGB}{9,54,150}
\definecolor{cubetop}{RGB}{40,112,238}
\definecolor{myblue}{RGB}{30,90,230}
\definecolor{myteal}{RGB}{0,160,160}
\definecolor{mygreen}{RGB}{20,150,40}
\definecolor{mypurple}{RGB}{110,50,160}
\definecolor{mybrown}{RGB}{90,40,10}
\definecolor{mygold}{RGB}{210,140,20}

\tikzset{
  ecPanel/.style={draw=black!65, rounded corners=1.2pt, line width=0.6pt, fill=white},
  ecSmallTitle/.style={font=\bfseries\large},
  ecSlot/.style={draw=black!70, rounded corners=2.5pt, minimum width=0.46cm, minimum height=3.05cm, fill=white},
  ecSlotBlue/.style={ecSlot, draw=editblue, line width=0.9pt},
  ecSlotRed/.style={ecSlot, draw=editred, line width=0.9pt},
  ecSuite/.style={draw=editblue, rounded corners=3pt, minimum width=3.8cm, minimum height=0.54cm, font=\footnotesize, text=editblue},
  ecModelBox/.style={draw=darkgreen, rounded corners=3pt, minimum width=4.2cm, minimum height=0.54cm, font=\footnotesize, align=center},
  suitePanel/.style={draw=black!70, line width=0.8pt, rounded corners=6pt},
  suiteDashedPanel/.style={draw=black!70, line width=0.8pt, dashed, rounded corners=6pt},
  suiteTitle/.style={font=\bfseries\large},
  suiteSubtitle/.style={font=\small\bfseries},
  filterbox/.style={draw=#1, fill=#1!10, rounded corners=3pt, align=center, font=\small\bfseries, inner sep=6pt, line width=0.8pt, minimum width=2.4cm, minimum height=1.1cm}
}

\newcommand{\hardtargetring}[2]{%
  \draw[white, line width=1.35pt, opacity=0.98] (#1,#2) circle[radius=0.46cm];
  \draw[red, line width=0.78pt, opacity=1] (#1,#2) circle[radius=0.46cm];
}

\newcommand{\ecSphere}[4]{\shade[ball color=#4] (#1,#2) circle (#3);}

\newcommand{\ecCubeObj}[3]{%
  \begin{scope}[shift={(#1,#2)}, scale=#3]
    \filldraw[fill=cubefront, draw=black, line width=0.45pt]
      (-0.25,-0.25) -- (0.11,-0.25) -- (0.11,0.11) -- (-0.25,0.11) -- cycle;
    \filldraw[fill=cuberight, draw=black, line width=0.45pt]
      (0.11,-0.25) -- (0.25,-0.11) -- (0.25,0.25) -- (0.11,0.11) -- cycle;
    \filldraw[fill=cubetop, draw=black, line width=0.45pt]
      (-0.25,0.11) -- (0.11,0.11) -- (0.25,0.25) -- (-0.11,0.25) -- cycle;
  \end{scope}%
}

\newcommand{\ecSlotDots}[4]{%
  \foreach \yy in {1.12,0.54,-0.04,-0.62} {\fill[#3] (#1,#2+\yy) circle (0.064);}
  \foreach \yy in {-0.92,-1.11,-1.30} {\fill[#4] (#1,#2+\yy) circle (0.032);}%
}

\newcommand{\ecModelGlyph}[3]{%
  \begin{scope}[shift={(#1,#2)}, scale=#3]
    \draw (0,0) -- (1.05,0.35) -- (1.05,1.85) -- (0,2.2) -- cycle;
    \node[font=\scriptsize] at (0.55,0.55) {model};
    \coordinate (c) at (0.58,1.25);
    \foreach \p in {(0.33,1.55), (0.82,1.55), (0.33,0.98), (0.82,0.98)} {
      \draw (c) -- \p; \fill[white] \p circle (0.065); \draw \p circle (0.065);
    }
    \fill[white] (c) circle (0.07); \draw (c) circle (0.07);
  \end{scope}%
}

\newcommand{\ecClipboardIcon}[2]{%
  \begin{scope}[shift={(#1,#2)}, scale=1.1]
    \draw[line width=0.6pt] (0,0) rectangle (0.26,0.43);
    \foreach \yy in {0.09,0.21,0.33} {\draw[line width=0.6pt] (0.07,\yy) -- (0.19,\yy);}
    \draw[line width=0.6pt] (0.08,0.46) -- (0.18,0.46);
  \end{scope}%
}

\newcommand{\ecNetworkIcon}[2]{%
  \begin{scope}[shift={(#1,#2)}, scale=0.65]
    \coordinate (c) at (0.5,0.6);
    \foreach \p in {(0.15,1.05), (0.85,1.05), (0.15,0.15), (0.85,0.15)} {
      \draw[line width=0.6pt] (c) -- \p; \fill[white] \p circle (0.09); \draw[line width=0.6pt] \p circle (0.09);
    }
    \fill[white] (c) circle (0.09); \draw[line width=0.6pt] (c) circle (0.09);
  \end{scope}%
}

\newcommand{\ecSearchIcon}[2]{%
  \begin{scope}[shift={(#1,#2)}, scale=1]
    \draw[line width=0.6pt] (0,0) rectangle (0.18,0.80);
    \draw[line width=0.6pt] (0.22,0.05) rectangle (0.40,0.40);
    \draw[line width=0.6pt] (0.52,0.17) circle (0.18);
    \draw[line width=0.8pt] (0.65,0.03) -- (0.88,-0.20);
  \end{scope}%
}

\newcommand{\suiteCube}[4]{%
  \begin{scope}[shift={(#1,#2)}, scale=#4]
    \filldraw[fill=#3!80!black, draw=black, line width=0.6pt] (-0.25,0) -- (0.25,0) -- (0.25,0.5) -- (-0.25,0.5) -- cycle;
    \filldraw[fill=#3!50!black, draw=black, line width=0.6pt] (0.25,0) -- (0.45,0.2) -- (0.45,0.7) -- (0.25,0.5) -- cycle;
    \filldraw[fill=#3!60, draw=black, line width=0.6pt] (-0.25,0.5) -- (0.25,0.5) -- (0.45,0.7) -- (-0.05,0.7) -- cycle;
  \end{scope}%
}

\newcommand{\suiteCyl}[4]{%
  \begin{scope}[shift={(#1,#2)}, scale=#4]
    \filldraw[fill=#3!80!black, draw=black, line width=0.6pt] (-0.25,0) arc (180:360:0.25 and 0.1) -- (0.25,0.8) arc (360:180:0.25 and 0.1) -- cycle;
    \filldraw[fill=#3!60, draw=black, line width=0.6pt] (0,0.8) ellipse (0.25 and 0.1);
  \end{scope}%
}

\newcommand{\EditCLEVRFigureOne}{%
\resizebox{0.98\textwidth}{!}{%
  \begin{tikzpicture}[x=1cm, y=1cm]

    \node[ecPanel, minimum width=6.6cm, minimum height=8.95cm, anchor=north west] at (0,14) {};
    \node[ecPanel, minimum width=6.6cm, minimum height=4.8cm, anchor=north west] at (0,4.85) {};
    \node[ecPanel, minimum width=7.4cm, minimum height=13.95cm, anchor=north west] at (6.8,14) {};
    \node[ecPanel, minimum width=10.2cm, minimum height=8.95cm, anchor=north west] at (14.4,14) {};
    \node[ecPanel, minimum width=5.0cm, minimum height=4.8cm, anchor=north west] at (14.4,4.85) {};
    \node[ecPanel, minimum width=5.0cm, minimum height=4.8cm, anchor=north west] at (19.6,4.85) {};

    \node[ecSmallTitle, anchor=west] at (0.25,13.55) {(a) Single atomic edit};
    \node[anchor=west, text=editblue, font=\bfseries] at (0.45,12.95) {BEFORE};
    \node[anchor=south west, inner sep=0pt] at (1.225,10.20)
      {\includegraphics[width=4.15cm,keepaspectratio]{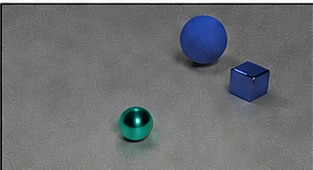}};

    \draw[editblue, dashed, rounded corners=4pt] (1.225,8.55) rectangle (5.375,9.95);
    \node[font=\small] at (3.3,9.45) {\textbf{Edit:}\quad sphere color};
    \node[font=\large] at (3.3,8.95) {\textcolor{editblue}{blue} $\to$ \textcolor{editred}{brown}};

    \node[anchor=west, text=editred, font=\bfseries] at (0.45,8.00) {AFTER};
    \node[anchor=south west, inner sep=0pt] at (1.225,5.25)
      {\includegraphics[width=4.15cm,keepaspectratio]{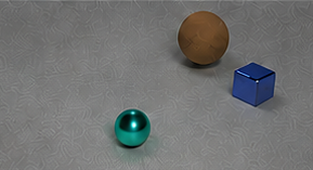}};

    \node[ecSmallTitle, anchor=west] at (0.25,4.45) {(b) Paired-scene benchmark};

    \node[anchor=center, inner sep=0pt] at (1.92,3.14)
      {\includegraphics[width=2.22cm,keepaspectratio]{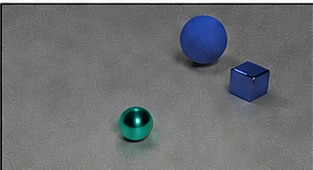}};
    \draw[editblue, rounded corners=2pt, line width=0.8pt] (0.78,2.51) rectangle (3.06,3.77);

    \node[anchor=center, inner sep=0pt] at (4.78,3.14)
      {\includegraphics[width=2.22cm,keepaspectratio]{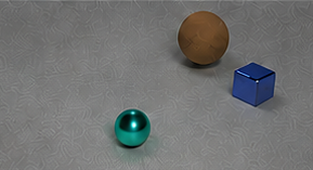}};
    \draw[editred, rounded corners=2pt, line width=0.8pt] (3.64,2.51) rectangle (5.92,3.77);

    \draw[Latex-Latex, dashed] (3.16,3.14) -- (3.52,3.14);
    \fill[darkgreen] (1.23,2.05) circle (0.18);
    \draw[white, line width=0.8pt] (1.14,2.05) -- (1.21,1.97) -- (1.34,2.15);
    \node[anchor=west, font=\small] at (1.62,2.05) {Same layout and object IDs};
    \fill[darkgreen] (1.23,1.40) circle (0.18);
    \draw[white, line width=0.8pt] (1.14,1.40) -- (1.21,1.32) -- (1.34,1.50);
    \node[anchor=west, font=\small] at (1.62,1.40) {Exactly one known edit};
    \fill[darkgreen] (1.23,0.75) circle (0.18);
    \draw[white, line width=0.8pt] (1.14,0.75) -- (1.21,0.67) -- (1.34,0.85);
    \node[anchor=west, font=\small] at (1.62,0.75) {Applied to one object};

    \node[ecSmallTitle, anchor=west] at (7.10,13.58) {(c) Object-centric representation};
    \node[font=\itshape\large] at (10.55,12.92) {per-object codes / slots};

    \node[text=editblue, font=\Large, anchor=west] at (7.10,12.08) {BEFORE};
    \ecSphere{9.62}{11.55}{0.30}{tealobj}
    \ecCubeObj{11.22}{11.55}{1.00}
    \ecSphere{12.82}{11.55}{0.30}{sphereblue}
    \ecModelGlyph{7.10}{8.10}{1.08}
    \draw[-Latex, line width=0.8pt] (8.45,9.30) -- (9.04,9.30);
    \node[ecSlot] at (9.62,9.30) {}; \ecSlotDots{9.62}{9.30}{black!75}{black!75}
    \node[ecSlot] at (11.22,9.30) {}; \ecSlotDots{11.22}{9.30}{black!75}{black!75}
    \node[ecSlotBlue] at (12.82,9.30) {}; \ecSlotDots{12.82}{9.30}{black!75}{black!75}

    \node[font=\huge] at (9.62,7.10) {$\approx$};
    \node[font=\huge] at (11.22,7.10) {$\approx$};
    \node[font=\huge, text=editred] at (12.82,7.10) {$\ne$};

    \node[text=editred, font=\Large, anchor=west] at (7.10,6.38) {AFTER};
    \ecSphere{9.62}{5.85}{0.30}{tealobj}
    \ecCubeObj{11.22}{5.85}{1.00}
    \ecSphere{12.82}{5.85}{0.30}{spherebrown}
    \ecModelGlyph{7.10}{2.40}{1.08}
    \draw[-Latex, line width=0.8pt] (8.45,3.60) -- (9.04,3.60);
    \node[ecSlot] at (9.62,3.60) {}; \ecSlotDots{9.62}{3.60}{black!75}{black!75}
    \node[ecSlot] at (11.22,3.60) {}; \ecSlotDots{11.22}{3.60}{black!75}{black!75}
    \node[ecSlotRed] at (12.82,3.60) {}; \ecSlotDots{12.82}{3.60}{editred}{black!75}

    \node[ecSmallTitle, anchor=west] at (14.75,13.55) {(d) What we measure};
    \ecSearchIcon{14.8}{12.15}
    \node[text=editblue, font=\bfseries, anchor=west] at (16.0,12.75) {Localization:};
    \node[anchor=west, font=\small] at (16.3,12.15) {$\bullet$\quad Edited-Object Accuracy (EOA)};
    \node[anchor=west, font=\small] at (16.3,11.55) {$\bullet$\quad Change Locality Score (CLS)};
    \node[text=editblue, font=\bfseries, anchor=west] at (14.80,10.35) {Stability:};
    \node[anchor=west, font=\small] at (15.35,9.80) {$\bullet$\quad No-Edit Drift (NED)};
    \node[text=editred, font=\bfseries, anchor=west] at (14.80,8.65) {Semantic faithfulness:};
    \node[anchor=west, font=\small] at (15.35,8.05) {$\bullet$\quad Target-Factor Accuracy (TFA)};
    \node[anchor=west, font=\small] at (15.35,7.45) {$\bullet$\quad Non-Target Preservation (NFP)};
    \node[anchor=west, font=\small] at (15.35,6.85) {$\bullet$\quad Unedited-Object Preservation (UOP)};
    \node[anchor=west, font=\small] at (15.35,6.25) {$\bullet$\quad Delta Scene-Graph Intervention Accuracy ($\Delta$SGIA)};
    \node[anchor=west, font=\small] at (15.35,5.65) {$\bullet$\quad Scene-Graph Intervention Accuracy (SGIA)};

    \ecClipboardIcon{15.45}{4.15}
    \node[font=\bfseries, anchor=west] at (15.85,4.40) {Suites};
    \node[ecSuite] at (16.9,3.55) {Atomic ID};
    \node[ecSuite] at (16.9,2.65) {No edit};
    \node[ecSuite] at (16.9,1.75) {Hard distractor};
    \node[ecSuite] at (16.9,0.85) {CoGenT OOD};
    \ecNetworkIcon{20.65}{4.00}
    \node[font=\bfseries, anchor=west] at (21.25,4.40) {Models};
    \node[ecModelBox] at (22.1,3.55) {GT-mask frozen backbones};
    \node[ecModelBox] at (22.1,2.65) {Native learned slots};
    \node[ecModelBox] at (22.1,1.75) {SAM 2 + frozen ViT};
    \node[ecModelBox] at (22.1,0.85) {\shortstack{Hybrid masks +\\ frozen pooling}};

  \end{tikzpicture}%
}%
}

\newcommand{\EditCLEVRFigureTwo}{%
\resizebox{0.84\columnwidth}{!}{%
\begin{tikzpicture}[x=1cm, y=1cm]
  \node[suitePanel, minimum width=14cm, minimum height=3.2cm, anchor=north west] at (0, 0) {};
  \node[suiteTitle] at (7, -0.5) {Atomic ID};

  \node[anchor=center, inner sep=0pt] at (3.5, -1.9) {\includegraphics[height=2.2cm,keepaspectratio]{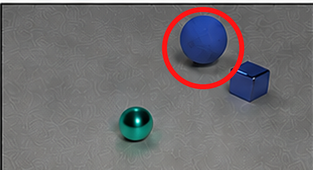}};
  \draw[-Latex, line width=2pt] (6.2, -1.9) -- (7.8, -1.9);
  \node[anchor=center, inner sep=0pt] at (10.5, -1.9) {\includegraphics[height=2.2cm,keepaspectratio]{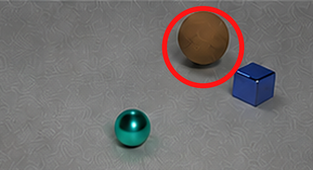}};

  \node[suitePanel, minimum width=14cm, minimum height=3.2cm, anchor=north west] at (0, -3.5) {};
  \node[suiteTitle] at (7, -4.0) {No edit};

  \node[anchor=center, inner sep=0pt] at (3.5, -5.4) {\includegraphics[height=2.2cm,keepaspectratio]{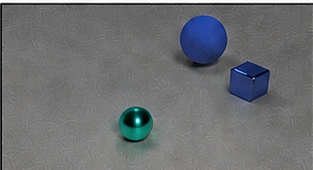}};
  \node[font=\Huge\bfseries] at (7, -5.4) {=};
  \node[anchor=center, inner sep=0pt] at (10.5, -5.4) {\includegraphics[height=2.2cm,keepaspectratio]{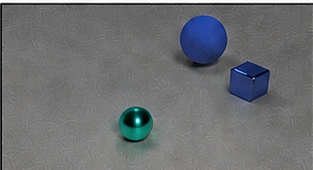}};

  \node[suitePanel, minimum width=14cm, minimum height=4.05cm, anchor=north west] at (0, -7.0) {};
  \node[suiteTitle] at (7, -7.5) {Hard distractor};

  \node[suiteSubtitle] at (3.5, -8.0) {Before};
  \node[suiteSubtitle] at (10.5, -8.0) {After};

  \node[anchor=center, inner sep=0pt] at (3.5, -9.05) {\includegraphics[width=3.8cm,keepaspectratio]{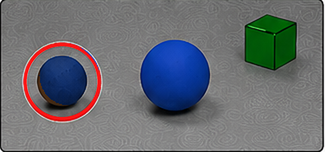}};
  \begin{scope}[shift={(3.5, -9.05)}]
    \hardtargetring{-1.15}{-0.05}
  \end{scope}
  \node[text=red, font=\footnotesize\bfseries] (targ) at (1.4, -10.25) {target};
  \draw[-Latex, red, thick] (targ) -- (2.2, -9.45);

  \node[font=\footnotesize\bfseries, align=center] (dist) at (5.4, -10.48) {distractor\\[-0.05em]\textit{\scriptsize(shares color + shape)}};
  \draw[-Latex, thick] (dist.north) -- (4.0, -9.45);

  \draw[-Latex, line width=2pt] (6.2, -9.05) -- (7.8, -9.05);
  \node[anchor=center, inner sep=0pt] at (10.5, -9.05) {\includegraphics[width=3.8cm,keepaspectratio]{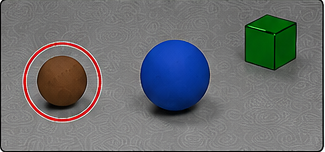}};
  \begin{scope}[shift={(10.5, -9.05)}]
    \hardtargetring{-1.15}{-0.05}
  \end{scope}

  \node[suitePanel, minimum width=14cm, minimum height=3.2cm, anchor=north west] at (0, -11.35) {};
  \node[suiteTitle] at (7, -11.85) {CoGenT OOD};

  \node[suiteSubtitle] at (3.5, -12.35) {Train (Condition A)};
  \suiteCube{2.6}{-13.85}{myblue}{1.3}
  \suiteCyl{4.4}{-13.85}{myteal}{1.3}

  \node[font=\bfseries] at (7, -12.35) {Swap};
  \begin{scope}[shift={(7,-13.55)}, scale=0.45]
    \draw[-Latex, thick, line width=1.2pt] (1.7,0.8) arc (30:150:1.9 and 1.2);
    \draw[-Latex, thick, line width=1.2pt] (-1.7,-0.8) arc (210:330:1.9 and 1.2);
  \end{scope}

  \node[suiteSubtitle] at (10.5, -12.35) {Test (Condition B)};
  \suiteCube{9.6}{-13.85}{myteal}{1.3}
  \suiteCyl{11.4}{-13.85}{myblue}{1.3}

  \node[suiteDashedPanel, minimum width=14cm, minimum height=4.2cm, anchor=north west] at (0, -14.85) {};
  \node[suiteTitle] at (7, -15.35) {CoGenT OOD Core};

  \node[font=\scriptsize\bfseries, align=center] at (1.1, -15.29) {CoGenT\\OOD};
  \draw[-Latex, thick, line width=1.2pt] (1.1, -15.61) -- (1.1, -16.27);

  \begin{scope}
    \clip (0.6, -16.45) -- (0.9, -17.15) -- (0.9, -18.05) arc (180:360:0.2 and 0.1) -- (1.3, -17.15) -- (1.6, -16.45) -- cycle;
    \fill[black!3] (0, -15.45) rectangle (2, -18.45);
  \end{scope}
  \draw[thick] (0.6, -16.45) -- (0.9, -17.15) -- (0.9, -18.05) arc (180:360:0.2 and 0.1) -- (1.3, -17.15) -- (1.6, -16.45);
  \draw[thick, fill=white] (1.1, -16.45) ellipse (0.5 and 0.15);
  \node[font=\scriptsize\bfseries, align=center] at (1.1, -18.63) {OOD-Core\\filter};

  \node[filterbox=myteal] (kcube) at (3.5, -16.65) {cube\\targets};
  \node[filterbox=mygreen] (kcyl) at (3.5, -18.05) {cylinder\\targets};

  \draw[-Latex, thick, line width=1.2pt, shorten >=0.4cm] (1.5, -17.35) -- (kcube.west);
  \draw[-Latex, thick, line width=1.2pt, shorten >=0.4cm] (1.5, -17.35) -- (kcyl.west);

  \node[draw=black!70, circle, thick, inner sep=2.5pt] at (5.2, -16.65) {\textcolor{mygreen}{\checkmark}};
  \node[draw=black!70, circle, thick, inner sep=2.5pt] at (5.2, -18.05) {\textcolor{mygreen}{\checkmark}};

  \draw[thick, dotted, black!50] (5.7, -15.85) -- (5.7, -18.65);
  \draw[thick, dotted, black!50] (9.6, -16.25) -- (9.6, -18.65);

  \node[suiteSubtitle] at (9.6, -15.85) {Allowed edits only};
  \node[font=\bfseries] at (7.65, -16.45) {Color edits};
  \node[font=\bfseries] at (11.8, -16.45) {Shape edits};

  \suiteCube{6.6}{-17.55}{myblue}{0.75}
  \draw[-Latex, thick] (7.35, -17.25) -- (7.95, -17.25);
  \suiteCube{8.7}{-17.55}{mygreen}{0.75}

  \suiteCyl{6.6}{-18.65}{myteal}{0.75}
  \draw[-Latex, thick] (7.35, -18.25) -- (7.95, -18.25);
  \suiteCyl{8.7}{-18.65}{mypurple}{0.75}

  \suiteCube{10.7}{-17.55}{myblue}{0.75}
  \draw[-Latex, thick] (11.45, -17.25) -- (12.05, -17.25);
  \suiteCyl{12.8}{-17.55}{myblue}{0.75}

  \suiteCyl{10.7}{-18.65}{myteal}{0.75}
  \draw[-Latex, thick] (11.45, -18.25) -- (12.05, -18.25);
  \suiteCube{12.8}{-18.65}{myteal}{0.75}
\end{tikzpicture}%
}%
}

\icmltitlerunning{EditCLEVR: Paired-Scene Interventions for Object-Centric Representations}

\begin{document}

\twocolumn[
\icmltitle{EditCLEVR: A Paired-Scene Intervention Benchmark for \\
           Compositional Faithfulness of Object-Centric Representations}

\icmlsetsymbol{equal}{*}
\begin{icmlauthorlist}
\icmlauthor{Anuraag Gadehothur Karnam}{lths,equal}
\icmlauthor{Tarunesh Sathish}{wghs,equal}
\end{icmlauthorlist}

\icmlaffiliation{lths}{Lebanon Trail High School}
\icmlaffiliation{wghs}{Walnut Grove High School}
\icmlcorrespondingauthor{Anuraag Gadehothur Karnam}{anuraag.gadehothurkarnam.761@k12.friscoisd.org}
\icmlcorrespondingauthor{Tarunesh Sathish}{sathitar000@k12.prosper-isd.net}

\icmlkeywords{object-centric learning, compositional generalization, intervention, benchmark}

\vskip 0.3in
]

\printAffiliationsAndNotice{\textsuperscript{*}These authors contributed equally. }

\begin{abstract}
Object-centric learning aims to represent scenes as objects whose properties can be reused in new combinations. Existing evaluations usually score segmentation, single-image factor prediction, or downstream accuracy, but these tests do not directly ask whether a per-object representation behaves correctly under a controlled semantic edit. We introduce \textbf{EditCLEVR}, a paired-scene intervention benchmark in which each example contains a before/after pair of CLEVR-style renders with the same object indices and scene layout, and either exactly one known attribute change on one known object or a no-edit re-render for drift measurement. The protocol includes probe-free diagnostics for representation-change localization and stability, together with probe-decoded semantic faithfulness metrics that test whether the predicted scene change matches the intended intervention across in-distribution and compositional out-of-distribution (OOD) suites, allowing code-space movement and decoded object-attribute correctness to be evaluated separately. We introduce the semantic metric Scene-Graph Intervention Accuracy (SGIA), which requires the full after-scene prediction to be correct and the only predicted before-to-after semantic change to be the intended object-factor edit. We also establish $\Delta$SGIA as a companion diagnostic that checks the single-site change pattern without requiring the full after-scene graph to be correct. Baseline evaluations on ground-truth-mask backbones, learned-slot models, SAM\,2 + frozen-ViT models, and one mask-feature hybrid indicate that CoGenT-OOD-core degradation can persist under ground-truth instance masks, that mask source accounts for part but not all of native performance, and that locality or stability alone can overstate semantic faithfulness. Code is available at \url{https://github.com/torux-bughunter/EditCLEVR}.
\end{abstract}

\begin{figure*}[t]
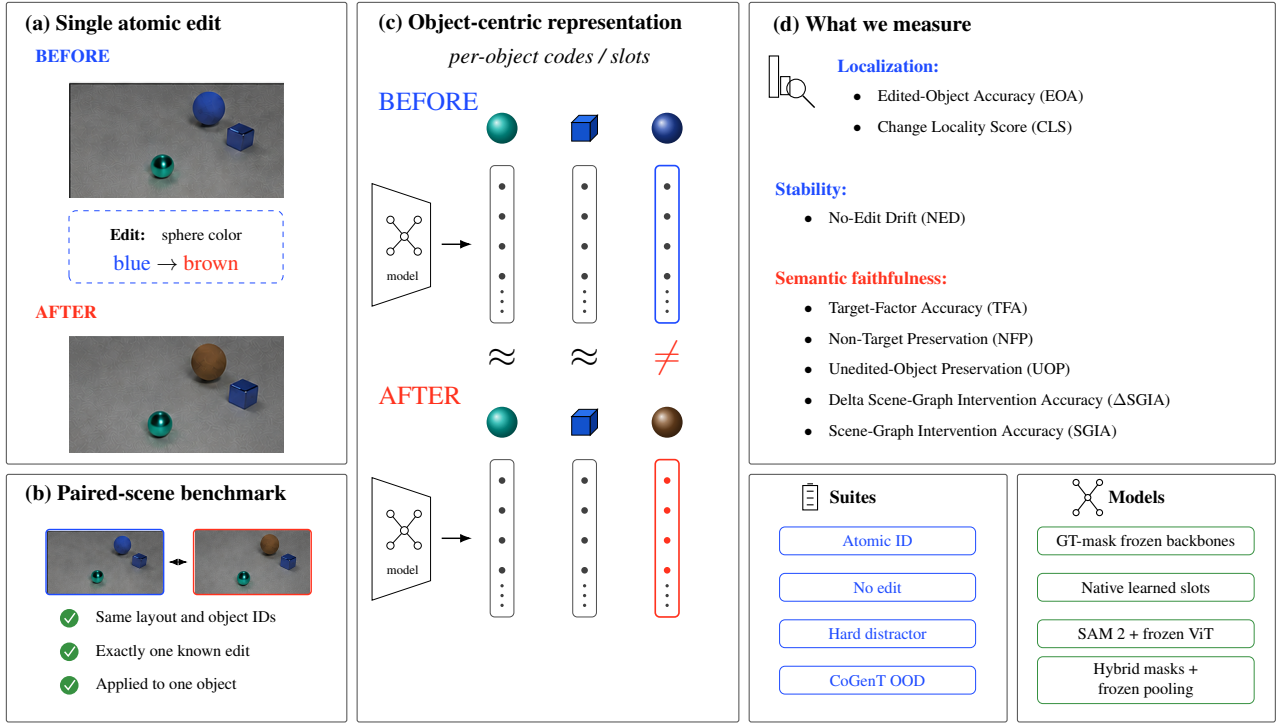

\centering
\EditCLEVRFigureOne
\caption{Evaluation protocol. EditCLEVR pairs a before scene with either a single known atomic edit or a no-edit re-render, extracts per-object representations, and evaluates whether the change localizes to the edited object, leaves other objects stable, and decodes to the intended object-attribute update.}
\label{fig:protocol}
\end{figure*}

\section{Introduction}
\label{sec:intro}

Object-centric learning is motivated by the idea that visual scenes can be represented through objects and their attributes, so that familiar factors can be recombined in new scenes~\cite{lake2017building,greff2020binding,dittadi2022generalization}. This notion is fundamentally an intervention-level one: if a single object changes color, material, size, or shape, the corresponding object code should change while other objects and factors remain stable. However, common evaluations only partially test this behavior. Segmentation metrics ask whether objects are discovered, factor probes ask whether attributes are decodable in a single image, and downstream accuracy mixes representation quality with task-specific shortcuts. These tests typically do not directly verify that a representation carries the intended before/after semantic change.

EditCLEVR operationalizes this criterion as a paired-scene benchmark (\cref{fig:protocol}). Each evaluation unit contains two CLEVR-style renders with the same scene layout and object indices, instance masks, object attributes, and metadata specifying the edited object and factor. Because the intervention is known, the benchmark can examine whether representation movement localizes to the edited object, unedited objects remain stable, and the decoded object-attribute graph changes only in the intended way. This paired design makes semantic faithfulness observable rather than inferred from a single-image score.

The benchmark stresses both standard generalization and compositional out-of-distribution (OOD) transfer. Its suites cover in-distribution atomic edits, no-edit re-renders, hard-distractor cases with a similar object, and a CoGenT-style OOD split in which cube/cylinder color palettes swap between train and test~\cite{johnson2017clevr}. A derived OOD-core slice keeps only color or shape edits on cubes and cylinders, excluding sphere cases whose color palette is unrestricted in both conditions and can dilute the intended color--shape shift.

We evaluate each intervention in terms of representation movement, no-edit stability, and decoded semantic change. Change locality (measured by CLS) and no-edit drift (NED) are probe-free diagnostics. Target-factor accuracy (TFA), non-target preservation (NFP), and unedited-object preservation (UOP) test semantic components of the edit. SGIA then conjoins absolute after-scene correctness with the requirement that the only predicted before-to-after change is the intended object-factor edit. This definition is deliberately strict, so we report it alongside a companion metric $\Delta$SGIA, which removes the absolute after-scene correctness requirement and acts as a relaxed diagnostic that measures intervention consistency.

\paragraph{Contributions.}
EditCLEVR contributes: (i) a dataset of CLEVR-derived paired scenes with known object attributes, instance masks, and either one object-factor edit or a no-edit re-render; (ii) a metric protocol that reports representation-level localization and stability alongside probe-decoded semantic faithfulness, with SGIA as the strict object-attribute scene-graph measure and $\Delta$SGIA/component metrics as diagnostics; and (iii) a baseline study indicating that compositional OOD degradation persists with ground-truth masks and that locality or stability alone can overstate semantic faithfulness.

\paragraph{Related work.}
EditCLEVR connects object-centric evaluation with compositional generalization. Object-centric benchmarks commonly report discovery quality on CLEVR, Multi-dSprites, or MOVi scenes~\cite{johnson2017clevr,greff2020multiobject,greff2022kubric}, or evaluate downstream factor prediction after object discovery. Disentanglement evaluations also measure factor decodability in single images~\cite{higgins2017beta,eastwood2018framework}, while identifiability results caution that such scores should not be read as proof that factors have been recovered without assumptions or supervision~\cite{locatello2019challenging}. EditCLEVR therefore makes a before/after intervention the evaluation unit and treats probe-decoded semantics as an operational readout, not as an identifiability claim.

Paired before/after evaluation has also appeared in vision-language benchmarking. BEAF, for example, manipulates visual scenes and evaluates whether VLMs understand the resulting changes~\cite{moon2024beaf}. EditCLEVR targets a different object of study: per-object visual representations. EditCLEVR extends paired before/after evaluation by making each scene pair object-indexed and measurable at three levels: code-space movement, object-discovery/matching, and decoded object-attribute graph change. It therefore asks whether object codes localize the edited object, preserve unedited objects, and decode to a single intended object-attribute change. The compared models span Slot Attention~\cite{locatello2020slot}, DINOSAUR slots over frozen ViT features~\cite{seitzer2023dinosaur}, and SAM/SAM\,2 segmentation-prior pipelines~\cite{kirillov2023sam,ravi2024sam2}. CLEVR-CoGenT motivates the color--shape OOD shift~\cite{johnson2017clevr}, and recent object-centric work also studies zero-shot transfer across datasets~\cite{didolkar2024zeroshot}.

\section{The EditCLEVR Benchmark}
\label{sec:bench}

\paragraph{Pairs and ground truth.}
EditCLEVR evaluates pairs rather than isolated images. Each example is a pair $(I, I')$ of RGB renders sharing a scene layout with 3--6 objects under visibility and overlap constraints. The pair is accompanied by before/after instance masks, object attributes (color, material, size, and shape), and edit metadata identifying the edited object $j^\star$ and edited factor. In edit suites, exactly one attribute of exactly one object changes. In the no-edit suite, the after image is a re-render of the same semantics, object positions, and object attributes with a different renderer seed, so movement in representation space is treated as drift rather than an edit.

\par\smallskip
\noindent\begin{minipage}{\columnwidth}
\centering
\EditCLEVRFigureTwo
\vspace{-0.45em}

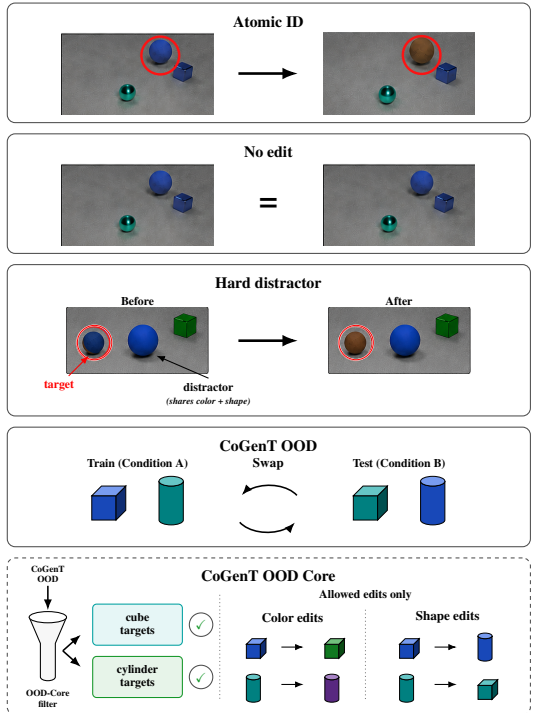
\captionof{figure}{EditCLEVR suite design. The suites stress in-distribution semantic edits, no-edit stability, instance disambiguation, and compositional OOD transfer. The OOD-core slice restricts evaluation to color/shape edits on cube/cylinder targets and removes sphere cases because spheres use the full color palette in both CoGenT conditions.}
\label{fig:suites}
\end{minipage}
\par\smallskip

\paragraph{Splits and suites.}
EditCLEVR contains 20{,}000 paired scenes. We use 10{,}000 training pairs and 1{,}000 validation pairs for probe fitting, then evaluate on four held-out suites: 3{,}000 atomic-ID pairs for in-distribution edits, 2{,}000 no-edit pairs for measuring drift under semantic no-ops, 2{,}000 hard-distractor pairs for target/distractor separation, and 2{,}000 CoGenT-OOD pairs for condition-B color-shape transfer. We also report a derived CoGenT-OOD-core slice containing only color or shape edits on cubes or cylinders before and after the edit ($n = 327$). This slice removes sphere cases, since spheres use the full color palette in both CoGenT conditions and can weaken the intended color-shape shift. The generated splits are balanced over edit factors and object counts before this OOD-core filtering step. The hard-distractor suite requires a visually similar object in the scene, but it is not meant to define a calibrated perceptual-difficulty scale; we use it as an instance-disambiguation stress test.

\section{Metrics}
\label{sec:metrics}

For each pair, let $T$ be the trusted object set, $j^\star$ the edited object, $f^\star$ the edited factor, $z_j,z'_j$ the L\textsubscript{2}-normalized object vectors, and $d_j=\lVert z_j-z'_j\rVert_2$. Ground-truth-mask rows use all ground-truth objects; predicted-object rows use the assignment protocol in \cref{sec:models}, with additional matching gates for semantic metrics. Localization and stability are defined directly in object-code space:
\begin{align*}
\mathrm{EOA} &= \mathbf{1}\!\left[\arg\max_{j\in T} d_j = j^\star\right], \\
\mathrm{CLS} &= \frac{d_{j^\star}}{\sum_{j\in T} d_j}, \\
\mathrm{NED} &= |T|^{-1}\sum_{j\in T} d_j
\quad\text{on no-edit re-renders.}
\end{align*}

For semantic metrics, let $y_{j,f},y'_{j,f}$ be ground-truth attributes and $\hat{y}_{j,f},\hat{y}'_{j,f}$ the probe predictions. We compute these metrics on edit suites; no-edit rows are used for NED. TFA requires the predicted after value of the edited factor to be correct,
\[
\hat{y}'_{j^\star,f^\star}=y'_{j^\star,f^\star}.
\]
NFP requires the edited object's non-target factors to be unchanged in prediction space, $\hat{y}'_{j^\star,f}=\hat{y}_{j^\star,f}$ for all $f\neq f^\star$. UOP requires the predictions for every non-edited object to be unchanged, $\hat{y}'_{j,f}=\hat{y}_{j,f}$ for all $j\neq j^\star$ in $T$. Thus TFA checks whether the intended after attribute is decoded, while NFP and UOP test whether the decoded change leaks to other factors or objects.

The single-site condition $\Delta_{\mathrm{site}}$ holds when the edited factor prediction changes at $(j^\star,f^\star)$ and no other predicted attribute in $T$ changes. We set
\[
\begin{array}{@{}l@{\;}c@{\;}l@{}}
\Delta\mathrm{SGIA} & = & \mathrm{TFA}\wedge\Delta_{\mathrm{site}},\\
\mathrm{SGIA} & = & \mathrm{SceneGraphExact}\wedge\Delta_{\mathrm{site}},
\end{array}
\]
where SceneGraphExact denotes after-frame object-attribute graph exactness: every decoded after-frame attribute in $T$ must equal ground truth, while the before frame enters through the predicted-change pattern. SGIA therefore conjoins two requirements by design--the after scene must decode correctly, and the predicted change pattern must be single-site--so it should be read together with $\Delta$SGIA and the component metrics. Because all semantic metrics are supervised-probe readouts, they test whether the representation supports a low-capacity decoder for the intervention semantics rather than proving that the latent representation is uniquely identifiable; Appendix~\ref{app:mlp} reports an MLP-probe replication to check probe sensitivity, following standard probe-interpretation cautions~\cite{hewitt2019designing}.

\begin{table*}[t]
\centering
\small
\setlength{\tabcolsep}{4.3pt}
\caption{Linear-probe results on EditCLEVR. ID is \textsc{atomic\_id} ($n=3{,}000$); OOD is \textsc{cogent\_ood\_core} ($n=327$); NED uses \textsc{no\_edit} ($n=2{,}000$). Values are row-level means; higher is better except NED. Native semantic metrics are conditional on the matching gate described in \cref{sec:models}. $\Delta$SGIA gap is OOD $\Delta$SGIA minus ID $\Delta$SGIA. Best overall values are bold; best native values are underlined.}
\label{tab:main}
\resizebox{\textwidth}{!}{%
\begin{tabular}{lcccccccccc}
\toprule
& \multicolumn{4}{c}{\textsc{atomic\_id} (ID)} & \multicolumn{4}{c}{\textsc{cogent\_ood\_core} (OOD)} & & \\
\cmidrule(lr){2-5}\cmidrule(lr){6-9}
Model & SGIA & $\Delta$SGIA & EOA & CLS & SGIA & $\Delta$SGIA & EOA & TFA & NED $\downarrow$ & $\Delta$SGIA gap \\
\midrule
\multicolumn{11}{l}{\emph{Ground-truth-mask frozen backbones}} \\
\quad DINO ViT-S/8 & 0.817 & 0.856 & \best{0.905} & 0.386 & \best{0.135} & 0.310 & \best{0.898} & 0.787 & 0.240 & $-0.546$ \\
\quad DINOv2 ViT-B/14 & 0.634 & 0.749 & 0.833 & 0.381 & 0.044 & 0.117 & 0.681 & 0.310 & 0.230 & $-0.632$ \\
\quad SigLIP\,2 ViT-B/16 & \best{0.862} & \best{0.884} & 0.841 & 0.377 & 0.123 & \best{0.398} & 0.877 & \best{0.980} & 0.270 & $-0.486$ \\
\midrule
\multicolumn{11}{l}{\emph{Native learned-slot discovery}} \\
\quad SA (conv) & 0.011 & 0.125 & 0.573 & \bothbest{0.437} & 0.000 & 0.163 & 0.594 & 0.863 & 0.073 & \bothbest{$+0.038$} \\
\quad DINOSAUR & \nativebest{0.640} & \nativebest{0.694} & \nativebest{0.799} & 0.374 & 0.021 & 0.137 & 0.696 & 0.395 & 0.116 & $-0.556$ \\
\midrule
\multicolumn{11}{l}{\emph{Native SAM\,2 proposals + frozen ViT}} \\
\quad SAM\,2 + DINO-S/8 & 0.485 & 0.560 & 0.713 & 0.360 & \nativebest{0.111} & \nativebest{0.237} & 0.702 & 0.798 & 0.264 & $-0.324$ \\
\quad SAM\,2 + DINOv2 & 0.557 & 0.631 & 0.762 & 0.363 & 0.061 & 0.155 & 0.640 & 0.406 & 0.248 & $-0.476$ \\
\quad SAM\,2 + SigLIP\,2 & 0.512 & 0.583 & 0.701 & 0.355 & 0.061 & 0.228 & \nativebest{0.719} & \nativebest{0.942} & 0.284 & $-0.355$ \\
\midrule
\multicolumn{11}{l}{\emph{Hybrid predicted masks + frozen pooling}} \\
\quad DINOSAUR-mask + DINO-S/8 & 0.112 & 0.210 & 0.331 & 0.236 & 0.009 & 0.035 & 0.333 & 0.547 & \bothbest{0.070} & $-0.175$ \\
\bottomrule
\end{tabular}
}%
\end{table*}

\section{Models and Evaluation Protocol}
\label{sec:models}

\paragraph{Model families as controls.}
We vary how object regions are obtained and which features are pooled. Ground-truth-mask rows pool frozen patch tokens under true instance masks, isolating representation quality when segmentation masks are controlled. These rows use DINO ViT-S/8 (384-d, 224 px)~\cite{caron2021dino}, DINOv2 ViT-B/14 (768-d)~\cite{oquab2023dinov2}, and SigLIP\,2 ViT-B/16 (768-d, 384 px)~\cite{tschannen2025siglip2}.

\paragraph{Native object discovery.}
The learned-slot rows test models that must discover objects: convolutional Slot Attention (SA)~\cite{locatello2020slot} and DINOSAUR, which applies Slot Attention to frozen DINO ViT-S/8 patch tokens with an MLP patch decoder~\cite{seitzer2023dinosaur}. The SAM\,2 rows use automatic proposals~\cite{kirillov2023sam,ravi2024sam2} and pool frozen ViT patch features inside those masks. The hybrid row uses DINOSAUR masks but frozen DINO-S/8 pooled features, giving a controlled mask-source comparison within one backbone family.

\paragraph{Native matching and gates.}
For native rows, we match predicted objects to ground-truth instances separately in each frame with a strict one-to-one best-overlap assignment; unused slots and extra SAM\,2 proposals are ignored. Semantic metrics require the edited object to have MatchBO $\geq 0.5$ in both frames, and UOP/SceneGraphExact use objects assigned in both frames. These native semantic scores are therefore conditional on matched objects, not full end-to-end discovery-plus-faithfulness scores. Appendix~\ref{app:disc} reports FG-ARI, MatchBO, MatchIoU, and the low-confidence MatchBO-exclusion rate; on \textsc{atomic\_id}, that exclusion rate is between $0.1\%$ and $1.4\%$ across native rows. A soft IoU-mixture alternative appears in Appendix~\ref{app:soft}.

\section{Results}
\label{sec:results}

The results in \cref{tab:main} highlight three empirical patterns. First, the ID--OOD-core drop persists in ground-truth-mask rows: DINO, DINOv2, and SigLIP\,2 backbones all lose most of their strict SGIA on \textsc{cogent\_ood\_core}, so the drop is not explained solely by object discovery. Second, mask source accounts for part, but not all, of the native-row gap: within the DINO-S/8 family, ground-truth masks give ID $\Delta$SGIA $0.856$, SAM\,2 masks give $0.560$, and DINOSAUR masks with frozen pooling give $0.210$, while DINOSAUR's learned slot features reach $0.694$. Third, locality and stability can overstate semantic faithfulness: the DINOSAUR-mask hybrid has the lowest NED ($0.070$) but weak SGIA, and SA has OOD-core TFA $0.863$ with strict SGIA $0.000$. Appendix~\ref{app:suites} reports all edit suites, Appendix~\ref{app:breakdowns} gives factor/object-count views, and Appendix~\ref{app:gates} details the SGIA conjunction.

\section{Discussion and Conclusion}
\label{sec:disc}

EditCLEVR illustrates how paired interventions complement segmentation and single-image decoding benchmarks. A model can discover objects, decode the edited factor, or remain stable under re-rendering noise while still changing the wrong parts of the predicted object-attribute graph. SGIA rewards only rows where preservation, after-scene correctness, and the single-site change pattern all hold, whereas $\Delta$SGIA and the component metrics help identify which part of the conjunction failed. The results suggest two related sources of error in these baselines. First, even with ground-truth instance masks, models still struggle under the CoGenT-derived OOD shift. Second, the choice of mask source affects native-model performance, but differences in masks alone do not explain the full performance gap.

\paragraph{Limitations and future work.}
EditCLEVR is synthetic and restricted to CLEVR-derived paired scenes, four discrete attributes, and single-object edits. The baselines cover ground-truth masks, learned slots, SAM\,2 proposals, frozen ViTs, and one hybrid, but not all object-centric, generative, or VLM-based systems. The semantic protocol uses supervised probes. Appendix~\ref{app:mlp} repeats the semantic evaluation with an MLP probe to check sensitivity to probe capacity. This ablation changes the readout, while keeping the paired scenes, edit labels, and metric definitions fixed. Natural images, relational or continuous edits, simultaneous edits, stronger discovery sources, calibration baselines, and factor-structured readouts are natural next steps. The accompanying artifacts support use of EditCLEVR through a Blender-backed dataset generator, dataset download and evaluation tools, a reference ground-truth-mask encoder, and a 20k-pair dataset with instance masks, object attributes, edit metadata, difficulty tags, split metadata, and no-edit re-render controls.

\section*{Impact Statement}

EditCLEVR is a synthetic diagnostic benchmark. Its direct societal impact is limited, and it should not be read as evidence of real-world visual robustness; controlled scenes and supervised probes complement, but do not replace, natural-image and safety-critical evaluations.

\newpage
\bibliographystyle{icml2026}
\bibliography{paper_safe}

\newpage
\appendix
\onecolumn

\section{Dataset and Implementation Details}
\label{app:impl}

\paragraph{Dataset.}
We render 20{,}000 paired scenes with a CLEVR-derived generator: $3$--$6$ objects per scene, balanced over the four edit factors, and constrained by visibility and overlap floors. Splits are \texttt{train}/\texttt{val}/\texttt{test\_id} (10k/1k/3k, suite \textsc{atomic\_id}), \texttt{test\_noop} (2k, \textsc{no\_edit}), \texttt{test\_hard} (2k, \textsc{hard\_distractor}), and \texttt{test\_cogent} (2k, \textsc{cogent\_ood}). The CoGenT regime swaps cube/cylinder color palettes between condition A (train/val/test\_id/test\_noop/test\_hard) and condition B (test\_cogent). The \textsc{cogent\_ood\_core} slice keeps rows whose edit factor is color or shape and whose edited object is a cube or cylinder both before and after the edit (327 rows). Each pair includes before/after RGB renders, before/after instance masks, before/after attribute records, edit metadata, and difficulty metadata. The hard-distractor split enforces a same-color and same-shape distractor, but it does not calibrate perceptual difficulty across examples; it should therefore be read as a controlled disambiguation stress test.

\paragraph{Backbones and discovery.}
DINO ViT-S/8 and DINOv2 ViT-B/14 run at $224\!\times\!224$; SigLIP\,2 ViT-B/16 runs at $384\!\times\!384$. Slot Attention~\cite{locatello2020slot} (denoted \textsc{SA}) uses a convolutional encoder at the dataset resolution with the slot count set above the maximum object count of the data; evaluation uses the same one-to-one best-overlap assignment as the other native rows. \textsc{DINOSAUR}~\cite{seitzer2023dinosaur} pools $28\!\times\!28$ frozen DINO ViT-S/8 patch tokens through Slot Attention with an MLP patch decoder; evaluation uses both predicted slot masks (for native rows and the hybrid) and slot features. \textsc{SAM\,2}~\cite{ravi2024sam2} runs in automatic-mask mode on each frame; per-mask proposals are then pooled with frozen ViT patch tokens and L\textsubscript{2}-normalized.

\paragraph{Object vectors and matching.}
All object vectors are L\textsubscript{2}-normalized in $\mathbb{R}^d$ (pooled patch tokens for ground-truth-mask rows; slot features for slot natives; mask-pooled patch tokens for SAM\,2 and the hybrid). Native rows use the strict one-to-one best-overlap assignment described in the main paper; each predicted object can be assigned to at most one ground-truth instance within a frame. Semantic metrics use a $\mathrm{MatchBO}\!\geq\!0.5$ gate on the edited object in both frames, while unconditioned variants keep the full denominator. The low-confidence rate in \cref{tab:disc} is the fraction of rows excluded by this gate on \textsc{atomic\_id}; it is small but nonzero ($0.1$--$1.4\%$ on \textsc{atomic\_id} across native rows), so native semantic scores should be interpreted as faithfulness on matched objects rather than as a complete discovery-plus-faithfulness measure. The soft IoU mixture in Appendix~\ref{app:soft} is an ablation, not the headline protocol.

\paragraph{Probes.}
The primary probe is one LogisticRegression per factor (color, material, size, shape) trained on train object vectors with seed-averaged accuracy. The MLP replication in Appendix~\ref{app:mlp} uses a 2-layer MLPClassifier ($512\times512$, Adam, early stopping, seed 42) for all rows. All metrics are computed per row; CIs are 95\% bootstrap intervals over per-row arrays.

\section{Full Per-Suite Linear Results}
\label{app:suites}

\Cref{tab:full-suites} reports SGIA / $\Delta$SGIA across the edit suites for every model. The hard-distractor suite remains close to \textsc{atomic\_id} for most rows, so it is best read as an instance-disambiguation diagnostic. The largest observed shift is the CoGenT OOD gap: every model loses strict SGIA on \textsc{cogent\_ood\_core}, and $\Delta$SGIA drops for every model except SA. That exception is not evidence of better faithfulness: SA's strict SGIA is essentially zero on OOD-core, so the relaxed score is capturing partial single-site change patterns under weak after-scene correctness.

\begin{table}[!htbp]
\AppendixTableSetup
\caption{Linear-probe SGIA / $\Delta$SGIA across all edit suites. NED is on \textsc{no\_edit}.}
\label{tab:full-suites}
\resizebox{\textwidth}{!}{%
\begin{tabular}{lccccccccc}
\toprule
& \multicolumn{2}{c}{\textsc{atomic\_id}} & \multicolumn{2}{c}{\textsc{hard\_distractor}} & \multicolumn{2}{c}{\textsc{cogent\_ood}} & \multicolumn{2}{c}{\textsc{cogent\_ood\_core}} & \\
\cmidrule(lr){2-3}\cmidrule(lr){4-5}\cmidrule(lr){6-7}\cmidrule(lr){8-9}
Model & SGIA & $\Delta$SGIA & SGIA & $\Delta$SGIA & SGIA & $\Delta$SGIA & SGIA & $\Delta$SGIA & NED $\downarrow$ \\
\midrule
DINO ViT-S/8 (GT-mask) & 0.817 & 0.856 & 0.823 & 0.866 & 0.142 & 0.387 & 0.135 & 0.310 & 0.240 \\
DINOv2 ViT-B/14 (GT-mask) & 0.634 & 0.749 & 0.647 & 0.749 & 0.036 & 0.252 & 0.044 & 0.117 & 0.230 \\
SigLIP\,2 ViT-B/16 (GT-mask) & 0.862 & 0.884 & 0.864 & 0.889 & 0.180 & 0.444 & 0.123 & 0.398 & 0.270 \\
SA (conv, native) & 0.011 & 0.125 & 0.016 & 0.144 & 0.001 & 0.091 & 0.000 & 0.163 & 0.073 \\
DINOSAUR (native) & 0.640 & 0.694 & 0.648 & 0.705 & 0.031 & 0.236 & 0.021 & 0.137 & 0.116 \\
SAM\,2 + DINO-S/8 & 0.485 & 0.560 & 0.488 & 0.571 & 0.136 & 0.273 & 0.111 & 0.237 & 0.264 \\
SAM\,2 + DINOv2 & 0.557 & 0.631 & 0.571 & 0.640 & 0.072 & 0.268 & 0.061 & 0.155 & 0.248 \\
SAM\,2 + SigLIP\,2 & 0.512 & 0.583 & 0.508 & 0.580 & 0.093 & 0.289 & 0.061 & 0.228 & 0.284 \\
DINOSAUR-mask + DINO-S/8 & 0.112 & 0.210 & 0.122 & 0.211 & 0.009 & 0.066 & 0.009 & 0.035 & 0.070 \\
\bottomrule
\end{tabular}
}%
\end{table}

\section{MLP Probe Replication}
\label{app:mlp}

\Cref{tab:mlp} reports the headline MLP values: SGIA, $\Delta$SGIA, and TFA for every model. The stronger decoder generally raises ID scores but does not remove the OOD failure mode: OOD-core SGIA remains low for every row, and high OOD-core TFA can still coexist with weak full-scene intervention faithfulness. We therefore keep the linear probe as the primary protocol and use the MLP run as a probe-sensitivity check.

\begin{table}[!htbp]
\AppendixTableSetup
\caption{Full MLP-probe headline results. ID is \textsc{atomic\_id}; OOD is \textsc{cogent\_ood\_core}.}
\label{tab:mlp}
\begin{tabular*}{\textwidth}{@{\extracolsep{\fill}}lcccccc@{}}
\toprule
& \multicolumn{3}{c}{\textsc{atomic\_id} (ID)} & \multicolumn{3}{c}{\textsc{cogent\_ood\_core} (OOD)} \\
\cmidrule(lr){2-4}\cmidrule(lr){5-7}
Model & SGIA & $\Delta$SGIA & TFA & SGIA & $\Delta$SGIA & TFA \\
\midrule
DINO ViT-S/8 (GT-mask) & 0.920 & 0.937 & 0.995 & 0.129 & 0.307 & 0.763 \\
DINOv2 ViT-B/14 (GT-mask) & 0.777 & 0.864 & 0.980 & 0.064 & 0.193 & 0.409 \\
SigLIP\,2 ViT-B/16 (GT-mask) & 0.930 & 0.944 & 0.996 & 0.021 & 0.450 & 0.985 \\
SA (conv, native) & 0.269 & 0.363 & 0.903 & 0.000 & 0.237 & 0.872 \\
DINOSAUR (native) & 0.737 & 0.773 & 0.984 & 0.088 & 0.225 & 0.550 \\
SAM\,2 + DINO-S/8 & 0.519 & 0.591 & 0.940 & 0.073 & 0.187 & 0.716 \\
SAM\,2 + DINOv2 & 0.474 & 0.546 & 0.935 & 0.070 & 0.172 & 0.482 \\
SAM\,2 + SigLIP\,2 & 0.529 & 0.602 & 0.937 & 0.035 & 0.228 & 0.939 \\
DINOSAUR-mask + DINO-S/8 & 0.508 & 0.555 & 0.970 & 0.070 & 0.161 & 0.737 \\
\bottomrule
\end{tabular*}
\end{table}

\section{Soft-Mixture Ablation for SAM\,2 Rows}
\label{app:soft}

The strict native protocol uses a one-to-one best-overlap assignment. As an ablation for the selected SAM\,2 rows, we also replace this with an IoU-weighted convex combination of SAM\,2 proposal features for each ground-truth object, $f_{\mathrm{obj}} = \sum_k w_k s_k$ with normalized IoU weights $w_k$. \Cref{tab:soft} reports that soft mixtures improve ID scores for both SAM\,2+DINO-S/8 and SAM\,2+SigLIP\,2, and they also improve OOD-core $\Delta$SGIA. The gains are not enough to remove the CoGenT gap, so the strict protocol remains the headline setting and the soft mixture is treated as a proposal-aggregation ablation.

\begin{table}[!htbp]
\AppendixTableSetup
\caption{Strict vs. soft IoU-mixture proposal aggregation for selected SAM\,2 rows (linear probes).}
\label{tab:soft}
\begin{tabular*}{\textwidth}{@{\extracolsep{\fill}}lcccccc@{}}
\toprule
& \multicolumn{2}{c}{\textsc{atomic\_id}} & \multicolumn{2}{c}{\textsc{cogent\_ood\_core}} & & \\
\cmidrule(lr){2-3}\cmidrule(lr){4-5}
Model & SGIA & $\Delta$SGIA & SGIA & $\Delta$SGIA & EOA(ID) & NED $\downarrow$ \\
\midrule
SAM\,2 + DINO-S/8 (strict) & 0.485 & 0.560 & 0.111 & 0.237 & 0.713 & 0.264 \\
SAM\,2 + DINO-S/8 (soft) & 0.635 & 0.700 & 0.132 & 0.292 & 0.810 & 0.243 \\
SAM\,2 + SigLIP\,2 (strict) & 0.512 & 0.583 & 0.061 & 0.228 & 0.701 & 0.284 \\
SAM\,2 + SigLIP\,2 (soft) & 0.678 & 0.734 & 0.073 & 0.301 & 0.769 & 0.270 \\
\bottomrule
\end{tabular*}
\end{table}

\section{Native Discovery Diagnostics}
\label{app:disc}

\Cref{tab:disc} reports the discovery-side diagnostics that gate the semantic metrics for native and hybrid rows: foreground ARI (FG-ARI), best-overlap coverage (MatchBO), matched-mask IoU (MatchIoU), and low-confidence rate. The low-confidence rate is the fraction of rows excluded by the edited-object MatchBO gate on \textsc{atomic\_id}. DINOSAUR masks have very high best-overlap but low IoU, meaning they reliably include the correct object while covering extra pixels. SAM\,2 masks have lower best-overlap than DINOSAUR masks but substantially higher IoU, which is consistent with their stronger frozen-feature results in the main-paper table. SA has the highest IoU in this table but a much larger low-confidence rate, so its semantic failures cannot be reduced to mask size alone.

\begin{table}[!htbp]
\AppendixTableSetup
\caption{Discovery diagnostics on \textsc{atomic\_id}. Higher is better except low-confidence rate. MatchBO is the best-overlap coverage score used for gates; MatchIoU is the IoU of the matched pair. Low-confidence is the MatchBO-gate exclusion rate for semantic metrics.}
\label{tab:disc}
\begin{tabular*}{\textwidth}{@{\extracolsep{\fill}}lcccc@{}}
\toprule
Model & FG-ARI & MatchBO & MatchIoU & LowConf $\downarrow$ \\
\midrule
SA (conv) & 0.869 & 0.912 & 0.420 & 0.014 \\
DINOSAUR & 0.975 & 0.983 & 0.098 & 0.002 \\
SAM\,2 + DINO/DINOv2/SigLIP\,2 & 0.913 & 0.974 & 0.270 & 0.001 \\
DINOSAUR-mask + DINO-S/8 & 0.975 & 0.983 & 0.098 & 0.002 \\
\bottomrule
\end{tabular*}
\end{table}

\section{Per-Factor and Object-Count Linear Breakdowns}
\label{app:breakdowns}

\Cref{tab:factor-count-breakdown} breaks the linear-probe results down by edited factor and by scene object count, pooling the three non-derived edit suites (\textsc{atomic\_id}, \textsc{hard\_distractor}, and \textsc{cogent\_ood}). Each cell reports SGIA/$\Delta$SGIA. The object-count columns show lower scores in larger scenes: strict SGIA falls from $0.738$ to $0.601$ for the SigLIP\,2 GT-mask, from $0.589$ to $0.295$ for DINOSAUR native, and from $0.527$ to $0.204$ for SAM\,2+SigLIP\,2. The factor columns show that failures are not isolated to a single factor; material and size are sometimes easier than color/shape for individual rows, but the OOD and object-count effects remain visible across the table.

\begin{table}[!htbp]
\AppendixTableSetup
\caption{Per-factor and per-object-count breakdowns for linear probes. Each cell is SGIA/$\Delta$SGIA, pooled over \textsc{atomic\_id}, \textsc{hard\_distractor}, and \textsc{cogent\_ood}; \textsc{no\_edit} and the derived OOD-core slice are excluded. These pooled values are descriptive because they mix ID, hard-distractor, and full CoGenT-OOD suites.}
\label{tab:factor-count-breakdown}
\begin{tabular*}{\textwidth}{@{\extracolsep{\fill}}lcccc@{}}
\toprule
& \multicolumn{4}{c}{Edited factor} \\
\cmidrule(lr){2-5}
Model & Color & Material & Size & Shape \\
\midrule
DINO ViT-S/8 (GT-mask) & 0.627/0.712 & 0.628/0.759 & 0.616/0.704 & 0.633/0.724 \\
DINOv2 ViT-B/14 (GT-mask) & 0.491/0.613 & 0.464/0.652 & 0.466/0.603 & 0.447/0.560 \\
SigLIP\,2 ViT-B/16 (GT-mask) & 0.682/0.770 & 0.658/0.764 & 0.646/0.726 & 0.684/0.779 \\
SA (conv, native) & 0.016/0.167 & 0.008/0.084 & 0.011/0.174 & 0.004/0.059 \\
DINOSAUR (native) & 0.471/0.545 & 0.470/0.604 & 0.471/0.567 & 0.463/0.548 \\
SAM\,2 + DINO-S/8 & 0.395/0.477 & 0.400/0.514 & 0.362/0.446 & 0.388/0.488 \\
SAM\,2 + DINOv2 & 0.442/0.535 & 0.430/0.572 & 0.417/0.517 & 0.402/0.495 \\
SAM\,2 + SigLIP\,2 & 0.414/0.513 & 0.400/0.520 & 0.359/0.460 & 0.393/0.499 \\
DINOSAUR-mask + DINO-S/8 & 0.095/0.182 & 0.082/0.174 & 0.085/0.155 & 0.079/0.165 \\
\bottomrule
\end{tabular*}
\vspace{0.35em}
\begin{tabular*}{\textwidth}{@{\extracolsep{\fill}}lcccc@{}}
\toprule
& \multicolumn{4}{c}{Objects in scene} \\
\cmidrule(lr){2-5}
Model & 3 & 4 & 5 & 6 \\
\midrule
DINO ViT-S/8 (GT-mask) & 0.700/0.786 & 0.633/0.729 & 0.617/0.717 & 0.554/0.666 \\
DINOv2 ViT-B/14 (GT-mask) & 0.561/0.673 & 0.495/0.631 & 0.452/0.592 & 0.361/0.532 \\
SigLIP\,2 ViT-B/16 (GT-mask) & 0.738/0.828 & 0.691/0.770 & 0.641/0.732 & 0.601/0.710 \\
SA (conv, native) & 0.030/0.229 & 0.008/0.159 & 0.001/0.075 & 0.001/0.022 \\
DINOSAUR (native) & 0.589/0.677 & 0.538/0.645 & 0.451/0.550 & 0.295/0.392 \\
SAM\,2 + DINO-S/8 & 0.525/0.599 & 0.455/0.543 & 0.371/0.474 & 0.194/0.309 \\
SAM\,2 + DINOv2 & 0.584/0.675 & 0.522/0.625 & 0.385/0.497 & 0.199/0.323 \\
SAM\,2 + SigLIP\,2 & 0.527/0.622 & 0.461/0.570 & 0.373/0.471 & 0.204/0.330 \\
DINOSAUR-mask + DINO-S/8 & 0.144/0.251 & 0.095/0.178 & 0.076/0.156 & 0.027/0.092 \\
\bottomrule
\end{tabular*}
\end{table}

\section{What Gets Counted as a Failure}
\label{app:gates}

SGIA is intentionally strict: a row passes only when the after-frame object-attribute graph is exactly correct on all trusted objects ($\mathrm{SceneGraphExact}=1$) \emph{and} the before-to-after prediction change pattern is confined to the edited object's edited factor (the $\Delta$SGIA single-site change pattern). This subsumes target-factor correctness, non-target preservation, and unedited-object preservation, but it additionally requires absolute after-scene correctness through SceneGraphExact, which is what makes it stricter than any factor-level conjunction. Component terms can look strong even when the joint event fails: SigLIP\,2 GT-mask TFA is $0.995$ on ID and $0.980$ on OOD-core, yet OOD-core SGIA is only $0.123$; SA has OOD-core TFA $0.863$ and $\Delta$SGIA $0.163$, yet SGIA is $0.000$. We therefore use SGIA as the headline strict measure, while using $\Delta$SGIA, TFA, NFP, UOP, the MLP check, and discovery diagnostics to separate decoding noise, intervention-pattern errors, probe effects, and object-assignment failures.

\section{Reproducibility}
\label{app:repro}

The accompanying code and dataset artifacts are available at \url{https://github.com/torux-bughunter/EditCLEVR}. The package includes Python tools for the Blender-backed CLEVR generator, dataset download, metric evaluation, the standardized \texttt{Evaluator}, and the reference ground-truth-mask encoder. The released dataset contains 20{,}000 paired examples, including no-edit re-render controls, with before/after RGB images, instance masks, scene JSON, object attributes, edit metadata, difficulty tags, and \texttt{splits.json}. The evaluation pipeline trains attribute probes on the train split and reports headline metrics with bootstrap confidence intervals, \texttt{results.json}, and per-split/suite CSVs.
\end{document}